%% file: root.tex
\documentclass[letterpaper, 10 pt, conference]{ieeeconf}  

\IEEEoverridecommandlockouts                              

\usepackage{graphics} 
\usepackage{epsfig} 
\usepackage{mathptmx} 
\usepackage{times} 
\usepackage{amsmath} 
\usepackage{amssymb}  
\usepackage{censor}
\usepackage{cite}
\usepackage{url}
\usepackage{booktabs}
\usepackage{multirow}
\usepackage{amssymb} 
\usepackage{makecell} 
\usepackage{colortbl}
\usepackage{makecell}
\usepackage{cuted}
\usepackage{capt-of}
\usepackage{algorithm}
\usepackage{booktabs}
\usepackage{multirow}
\usepackage{hyperref}
\usepackage{paralist}
\usepackage{algpseudocode} 
\title{\LARGE \bf
ForEnt: A Multi-Modal Dataset for Characterizing Quadruped Robot Entrapments in Forest Environments
}

\author{Natapat Kirdwichai and Danesh Tarapore
\thanks{The authors are with the Faculty of Engineering and Physical Sciences, University of Southampton, United Kingdom. Email: {\tt\small nk3g22@soton.ac.uk}%
}}

\begin{document}

\maketitle
\thispagestyle{empty}
\pagestyle{empty}

\begin{strip}
    \vspace{-1.75cm}
    \centering
    \includegraphics[width=\textwidth]{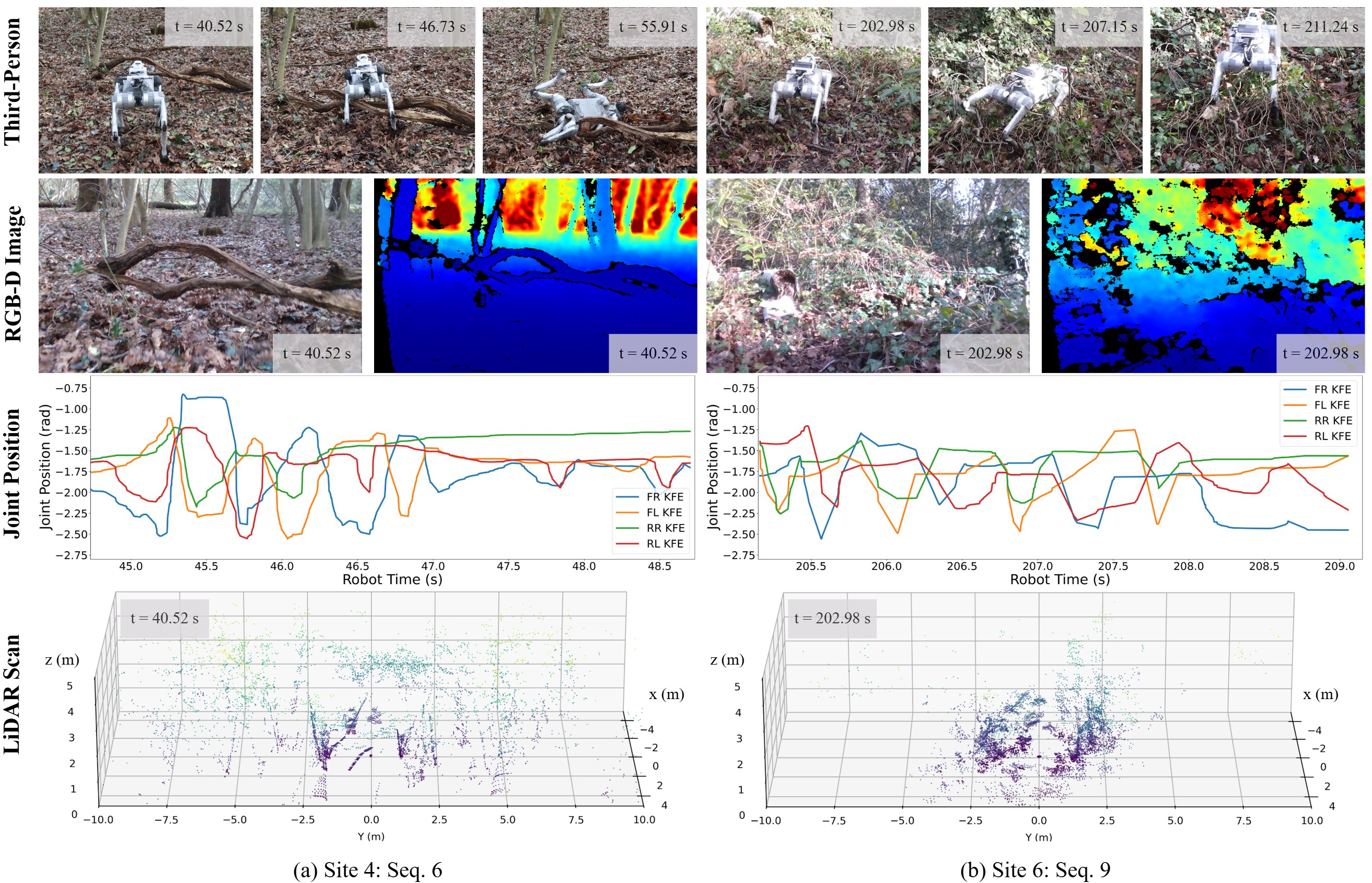}
    \captionof{figure}{Multi-modal sensor data recorded during Seq.~6 at Site~4 and Seq.~9 at Site~6, illustrating examples of the two most frequent entrapment types in the ForEnt dataset. (a) A trip event caused by a fallen log, resulting in a fall that required physical intervention. (b) A vine entanglement event, requiring the operator to reverse the robot via teleoperation. For both cases, the pre-entrapment exteroceptive data (RGB image, depth map, and single LiDAR scan) correspond to the timestamp of the leftmost third-person frame. In the depth maps, blue indicates proximal surfaces, red indicates distant surfaces, and black denotes invalid or missing sensor returns. In the LiDAR scans, brighter colors represent elevated points. The corresponding proprioceptive joint position data is plotted across a $\pm 2$~s window centered on the timestamp of the middle third-person frame, capturing the kinematic onset of the entrapment. This synchronized alignment of pre-failure visual context and high-frequency physical failure signatures provides the necessary foundation for evaluating both predictive traversability and reactive entrapment detection algorithms.}
    \label{fig:teaser}
\end{strip}

\begin{abstract}
Legged robots are increasingly deployed in forests for ecological surveying and monitoring, yet their autonomy is often interrupted consequent to the challenges posed in traversing forest environments. Forest entrapments, for example, when a robot’s legs are ensnared in vines or other vegetation, result in loss of stability and toppling. Such events not only disrupt the mission and require manual intervention, but also risk damage to the robot hardware. To address the absence of a dedicated dataset to investigate these failure modes in forest environments, we present ForEnt, a multi-modal dataset collected with the low-cost Unitree Go2 quadruped across eight forest sites in the Southampton Common Woodlands, UK. For our dataset, over approximately 1.7~km of traversals in 11 sequences were conducted, yielding 69 recorded entrapment events. ForEnt includes time-synchronized RGB-D images, LiDAR scans, proprioceptive data, and third-person video, enabling analysis of terrain factors contributing to entrapment and providing labeled sensor streams for reproducible benchmarking. By supporting the evaluation of entrapment detection strategies, ForEnt lowers the barrier to developing robust quadruped robot deployments in challenging forest environments.
\end{abstract}


\input{Sections/Introduction}

\input{Sections/Related_Work}
\input{Sections/Hardware_Setup}
\input{Sections/Dataset}
\input{Sections/Evaluation}

\input{Sections/Conclusion}






\bibliographystyle{IEEEtran}
\bibliography{references}

\end{document}

%% file: Sections/Introduction.tex
\section{Introduction}

Forest cover approximately 31\% of the Earth's land area \cite{UNDESA2021GlobalForestGoals} and play a critical role in carbon sequestration, biodiversity preservation, and climate resilience. Monitoring these vast, cluttered, and often inaccessible environments at scale remains logistically demanding using traditional manual surveys. In recent years, ground robots---including quadrupeds and wheeled rovers---have been increasingly deployed for automated forest monitoring tasks \cite{Pollayil2023RoboticMonitoringForests,Gromek2025WildfireUGVViewpoint}. These platforms enable rapid and repeatable acquisition of structural and radiometric forest parameters, including canopy cover, above-ground biomass  and tree-stem metrics such as diameter at breast height \cite{Mattamala2025BuildingForestInventories}. However, maintaining reliable autonomy in cluttered and vegetation-dense forest terrain remains an open challenge.



\definecolor{highlightgray}{gray}{0.9}

\begin{table*}[h]
\centering
\caption{Comparison of field robotics datasets. The table is grouped by platform type: Wheeled (top), Industrial Legged (middle), and Mid-Size/Low-Cost Legged (bottom).}
\label{tab:dataset_comparison}
\resizebox{\textwidth}{!}{%
\begin{tabular}{cccccc}
\toprule
\textbf{Dataset} & \textbf{Platform} & \textbf{Environment} & \textbf{External Sensors} & \textbf{Primary Focus} & \textbf{Failure Annotations} \\ 
\midrule

TartanDrive \cite{triest2022tartandrive} & ATV & Off-Road Trails / Grassland & RGB, IMU, CAN & Outdoor Interaction & $\checkmark$ \\
TartanDrive 2.0 \cite{sivaprakasam2024tartandrive}  & ATV & Off-Road Trails / Grassland & RGB, LiDAR & Outdoor Interaction & $\checkmark$ \\
GOOSE \cite{mortimer2024goose}  & Jackal & Off-Road Trails & RGB, LiDAR & Perception & $\times$ \\
\midrule

Nebula Multi \cite{chang2022lamp} & Spot & Subterranean / Caves & LiDAR, Thermal & SLAM & $\times$ \\
M3ED \cite{chaney2023m3ed} & Spot & Street / Forest & Event, RGB, LiDAR & Event VIO & $\times$ \\
Habitat 9210 \cite{Pollayil2023RoboticMonitoringForests}  & ANYmal & Forest & LiDAR, RGB & Biomass Estimation & $\times$ \\
GrandTour \cite{frey2026grandtour} & ANYmal & Industrial / Urban / Forest & LiDAR, RGB-D & Navigation & $\times$ \\
\midrule

CEAR \cite{zhu2024cear}  & Mini Cheetah & Campus / Street / Indoor & Event, RGB-D, LiDAR & Perception & $\times$ \\
DiTer / DiTer++ \cite{jeong2024diter,kim2025diterplusplus} & Go1/Go2 & Park / Street & RGB, Thermal, LiDAR & SLAM / Navigation & $\times$ \\
M-SEVIQ \cite{cao2026mseviq}  & Go2 & Park / Street / Indoor & Stereo Event, RGB & Perception & $\times$ \\

\textbf{ForEnt (Ours)} & \textbf{Go2} & \textbf{Forest} & \textbf{RGB-D, LiDAR} & \textbf{Failure Analysis} & \boldmath$\checkmark$ \\ 
\bottomrule
\end{tabular}
}
\end{table*}

Quadruped robots, in comparison to wheeled and tracked platforms, tend to negotiate unstructured terrain and maneuver across obstacles more reliably, particularly where vegetation, roots, deadwood, and uneven substrates limit traction and clearance for wheeled and tracked mobility \cite{Hutter2017ANYmalHarshEnv, 11246357}. However, on forest terrain, quadrupeds remain vulnerable to \emph{entrapment}, a condition in which the robot is unable to sustain forward progress without external assistance. Entrapment events arise from a combination of environmental factors and system limitations that degrade navigation in forest environments, where compliant vegetation like vines or low branches may catch legs in the swing phase and induce entanglement or destabilize the body \cite{Mattamala2025BuildingForestInventories, Yim2023EntanglementsIROS}. Moreover, weak or variable foot contact on boggy, waterlogged, or rocky ground increases the likelihood of slips and stalls \cite{Mattamala2025BuildingForestInventories, 9981660}. Environmental hazards are further exacerbated by sensor limitations that cause odometry drift over long traverses \cite{Mattamala2025BuildingForestInventories} and perception failures under occlusion and poor illumination \cite{Sathyamoorthy2023VERN, Weerakoon2024VAPOR}. Beyond interrupting autonomy, such events raise the risk of damage to the platform and its onboard sensors.




Despite the vulnerability of quadrupeds to entrapment in natural forest terrain, existing forest robotics datasets are primarily designed for ecological measurement tasks such as biomass estimation and forest inventory \cite{Chirici2023RoboticsForestInventories, Pollayil2023RoboticMonitoringForests, Mattamala2025BuildingForestInventories, Frey2025BoxiRSS}. Consequently, the data collection protocols in these datasets are optimized for successful and uninterrupted traversal. As a result, mobility failures are rarely captured or systematically annotated, limiting rigorous study of vegetation-induced entrapment and locomotion breakdown in natural forest terrain. A dedicated dataset that explicitly captures and labels such failure modes is therefore essential for evaluating the robustness of robot autonomy in forest environments. Such evaluation is critical for safe and scalable deployment of legged robots in long-duration field operations.

To address this gap, we present \emph{ForEnt}, a multi-modal forest entrapment dataset collected using a quadruped platform across eight distinct forest sites. The dataset comprises approximately 1.7~km of traversal and 69 manually annotated entrapment events, with time-synchronized RGB-D imagery, 3D LiDAR scans, and high-frequency proprioceptive measurements. Rather than prioritizing traversal distance, the ForEnt dataset is structured to concentrate and systematically annotate diverse real-world mobility breakdowns. This design contrasts with mission-centric forest datasets that emphasize nominal locomotion, and instead captures vegetation- and terrain-induced failures in natural forest environments. 

ForEnt's utility is demonstrated on two evaluation tasks:
\begin{itemize}
    \item Visual terrain traversability estimation under vegetation-dense forest conditions; and
    \item Proprioceptive entanglement detection in dense overhanging and underfoot vines.
\end{itemize}

\noindent Rather than proposing new algorithms, we evaluate representative existing pipelines to illustrate how realistic forest entrapment scenarios expose limitations that may not be apparent in existing datasets. By releasing these multi-modal, failure-centric sequences, we aim to enable the reproducible and systematic evaluation of forest autonomy in challenging natural terrain.

%% file: Sections/Related_Work.tex
\section{Related Work}
A number of datasets have been developed to support autonomous operation of wheeled and legged robots in outdoor environments (summarized in Table~\ref{tab:dataset_comparison}). Such field robotics datasets have enabled research in state estimation, perception, and terrain-aware navigation across a diverse range of structured and unstructured environments \cite{10876033}. 

In forests, datasets collected using quadruped robots have been employed to estimate the structural and radiometric characteristics of trees \cite{9981660} and support applications such as biomass estimation and forest inventory \cite{Chirici2023RoboticsForestInventories}. However, these datasets are typically collected under protocols designed to promote uninterrupted traversal and consistent environmental coverage with minimal disruptions, and consequently emphasize nominal locomotion. As a result, evaluation in these datasets is largely confined to near-nominal operation, providing limited insight into autonomy under mobility breakdown.




Failure-centric forest datasets are essential for evaluating autonomy under the conditions where robot mobility is most likely to degrade or fail \cite{11246357}. For autonomy, terrain traversability estimation methods rely on observing both traversable and non-traversable terrain interactions in order to learn or assess the limits of safe locomotion \cite{10876033}. Similarly, proprioceptive entrapment detection methods require representative examples of mobility impairment to distinguish nominal locomotion from failure states \cite{9981660}. However, existing approaches, including visual traversability estimation frameworks such as Wild Visual Navigation (WVN) \cite{frey23fast} and proprioceptive entanglement detection methods such as Momentum-Based Observer (MBO) \cite{Yim2023EntanglementsIROS}, have primarily been evaluated on datasets that do not systematically capture vegetation-induced entrapment or terrain-induced immobilization. In the absence of datasets that explicitly capture and annotate such events, evaluation is limited to nominal or near-nominal operation, and does not fully characterize performance under realistic failure conditions. 



The absence of failure-centric robot locomotion data in forests disproportionately impacts mid-sized quadrupeds---platforms typically weighing on the order of 10--20 kg---such as the Unitree Go2\footnote{\label{fn:go2}\url{https://www.unitree.com/go2}} and Mini Cheetah \cite{8793865}. These platforms lower the barrier to entry for large-scale multirobot forest monitoring \cite{10.3389/frobt.2020.00083}. However, their reduced ground clearance and lower mass increase susceptibility to entrapment by vines, undergrowth, and leaf litter compared to larger quadrupeds such as the Spot\footnote{\label{fn:spot}\url{https://bostondynamics.com/products/spot/}} and ANYmal \cite{Hutter2017ANYmalHarshEnv}. To date, existing datasets featuring mid-sized platforms---including CEAR \cite{zhu2024cear}, DiTer/DiTer++ \cite{jeong2024diter, kim2025diterplusplus}, and M-SEVIQ \cite{cao2026mseviq}---are predominantly recorded in semi-structured vegetated spaces (e.g., campus parks), structured urban (e.g., street pavements), or indoor environments (see Table~\ref{tab:dataset_comparison}). To the best of our knowledge, no dataset provides temporally aligned, type-specific annotation of vegetation-induced entrapment in natural forest terrain.


Our ForEnt dataset addresses this gap by providing time-synchronized LiDAR, RGB-D, and high-frequency proprioceptive data collected on a cost-effective mid-sized quadruped platform---the Unitree Go2---with systematically annotated entrapment intervals and severity labels. This enables reproducible benchmarking of traversability estimation and entanglement detection methods under realistic forest conditions encountered by deployable field robots. Importantly, while collected on a single platform, the observed failure mechanisms in ForEnt reflect terrain-robot interactions that are characteristic of mid-sized legged systems operating in cluttered forest environments.


%% file: Sections/Hardware_Setup.tex
\section{Experimental Setup}

\begin{table*}[t]
\centering
\caption{Overview of the Hardware Specifications Used in the ForEnt Dataset}
\label{table:sensor_specs}
\small
\renewcommand{\arraystretch}{1.1} 
\begin{tabular*}{\textwidth}{@{\extracolsep{\fill}} c c p{5.0cm} c p{2.2cm} p{2.8cm} @{}}
\toprule
\textbf{Hardware} & \textbf{Type} & \textbf{Characteristics} & \textbf{Rate} & \textbf{Topic name} & \textbf{Message type} \\ 
\midrule
\begin{tabular}[t]{@{}c@{}}3D LiDAR \\ \& IMU\end{tabular} & 
\begin{tabular}[t]{@{}c@{}}Livox Mid-360 \\ (ICM-40609)\end{tabular} & 
$360^\circ$ FOV, 40~m range. Non-repetitive scanning pattern. & 
10~Hz & 
/livox/lidar & 
livox\_ros\_driver2/ \newline msg/CustomMsg \\ 
& & 6-axis integrated IMU. & 200~Hz & /livox/imu & sensor\_msgs/ \newline msg/Imu \\ 
\midrule
\begin{tabular}[t]{@{}c@{}}RGB-D Camera \\ \& IMU\end{tabular} & 
\begin{tabular}[t]{@{}c@{}}Intel D435i \\ (BMI055)\end{tabular} & 
RGB: $640 \times 480$, $69^\circ$ HFOV. 8~ms fixed shutter, dynamic gain. & 
15~Hz & 
/camera/color/ \newline image\_raw & 
sensor\_msgs/ \newline msg/Image \\
& & Depth: Stereo IR, $85.2^\circ$ HFOV, $58^\circ$ VFOV, 0.3--3.0~m range. & 15~Hz & /camera/depth/ \newline image\_rect\_raw & sensor\_msgs/ \newline msg/Image \\
& & 6-axis integrated IMU. & 200~Hz & /camera/imu & sensor\_msgs/ \newline msg/Imu \\ 
\midrule
\begin{tabular}[t]{@{}c@{}}Robot \\ Platform\end{tabular} & 
Unitree Go2 & 
Joint encoders \& torques (12 actuators), 6-axis body IMU, foot contact force, battery voltage/current. & 
300~Hz & 
/lowstate & 
unitree\_go/ \newline msg/LowState \\
\bottomrule
\end{tabular*}
\vspace{-3mm}
\end{table*}

\begin{figure}
    \centering
    \includegraphics[width=\linewidth]{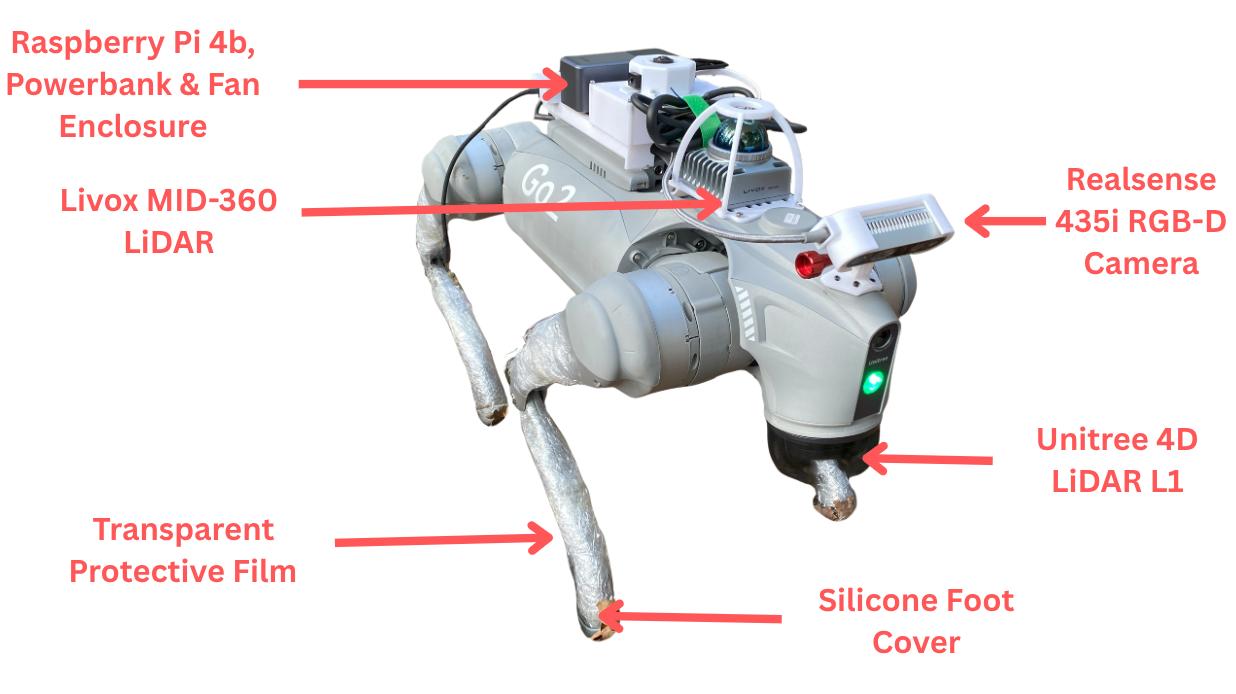}
    \caption{The Unitree Go2 Edu with its exteroceptive sensors, including the following: (A) A front-mounted RealSense D435i RGB-D camera; (B) An upward-mounted external Livox MID-360 LiDAR with a custom 3D printed cage; and (C) An onboard Unitree 4D LiDAR L1. Transparent protective film and silicone foot covers were added to reduce cosmetic wear from scratches and falls during forest trials.}
    \label{fig:Go2_extra_sensors}
    \vspace{-3mm}
\end{figure}

The Unitree Go2 quadruped (Go2) was used for the forest entrapment data collection (see Fig.~\ref{fig:Go2_extra_sensors}). The robot has 12 actuated degrees of freedom, with three joints per leg: Hip Abduction/Adduction (HAA), Hip Flexion/Extension (HFE), and Knee Flexion/Extension (KFE). The Edu variant features an onboard NVIDIA Jetson Orin Nano, which serves as the primary compute node for data aggregation and synchronization. All sensor streams, including high-frequency proprioception, LiDAR point clouds, and RGB-D data, were recorded into a single ros2bag per sequence. The custom sensors are rigidly mounted to the robot's torso using 3D-printed fixtures.

\subsection{Locomotion and Control Protocol}
To establish a consistent baseline for entrapment analysis, the robot was operated under a standardized locomotion protocol throughout all dataset sequences. The Go2 utilized the Unitree RL-Lab trotting gait\footnote{\url{https://github.com/unitreerobotics/unitree_rl_lab}}, maintaining a default nominal body height of 0.3~m and a fixed gait frequency of 2~Hz. The maximum commanded forward velocity was 1~m/s, and raw joystick inputs were recorded within the dataset to enable the estimation of instantaneous commanded velocities. Fixing these parameters ensures that the dataset isolates the natural terrain susceptibility of the Go2 without the confounding effects of gait adaptation.

During collection, a teleoperator guided the robot along linear or serpentine trajectories to simulate forest surveying tasks \cite{Chirici2023RoboticsForestInventories, Mattamala2025BuildingForestInventories}. To accurately capture realistic failure signatures, the operator adhered to a consistent intervention policy: if the robot became stuck, the operator maintained the forward command and only applied corrective inputs after approximately 3~seconds of continuous entrapment. If the robot remained immobilized for 10-15~seconds or suffered a fall, it was manually repositioned to adjacent flat ground along the path to resume the trial.

\subsection{Sensor Overview}
To support the evaluation of terrain traversability and proprioceptive entanglement detection algorithms, the Go2 is equipped with an Intel RealSense D435i RGB-D camera and a Livox Mid-360 LiDAR (sensor specifications detailed in Table~\ref{table:sensor_specs}). The front-facing RGB camera is utilized to extract semantic terrain information, while the Livox LiDAR employs a non-repetitive scanning pattern to generate dense 3D point clouds to support geometric traversability estimation, environmental mapping, and LiDAR-inertial odometry. Furthermore, kinematics odometry from the Unitree SDK is recorded to provide an independent comparative baseline for state estimation during mobility breakdowns.

\subsection{Calibration}
Intrinsic and extrinsic calibrations were conducted to ensure multi-modal consistency, where calibration parameters are included in the dataset.
\subsubsection{Intrinsic Calibration}
Intrinsic parameters for the D435i were estimated using the Kalibr toolbox\footnote{\url{https://github.com/ethz-asl/kalibr}} to determine focal length, principal point, and distortion coefficients. The calibration achieved a mean projection error of 0.66~px. For the IMU, noise densities and random walk parameters were characterized using the Allan Variance ROS toolbox\footnote{\url{https://github.com/ori-drs/allan_variance_ros}} based on a 4-hour static recording. 
\subsubsection{Extrinsic Calibration}
Spatial transformations between the LiDAR, D435i, and robot base were determined using CAD measurements and refined via the Kalibr toolbox. We validated the extrinsics by checking the alignment between the projected LiDAR point clouds and the RGB-D frames.

\subsection{Time Synchronization}
The D435i and Livox LiDAR shared the system clock of the onboard Jetson Orin Nano. Timestamps reflect the capture time at the sensor hardware level. To synchronize the third-person camera, a ``sitting-to-standing'' motion was performed at the start and end of each sequence, allowing for time alignment with the sensor data during post-processing.


%% file: Sections/Dataset.tex
\section{Dataset}

\begin{figure*}
    \centering
    \includegraphics[width=\textwidth]{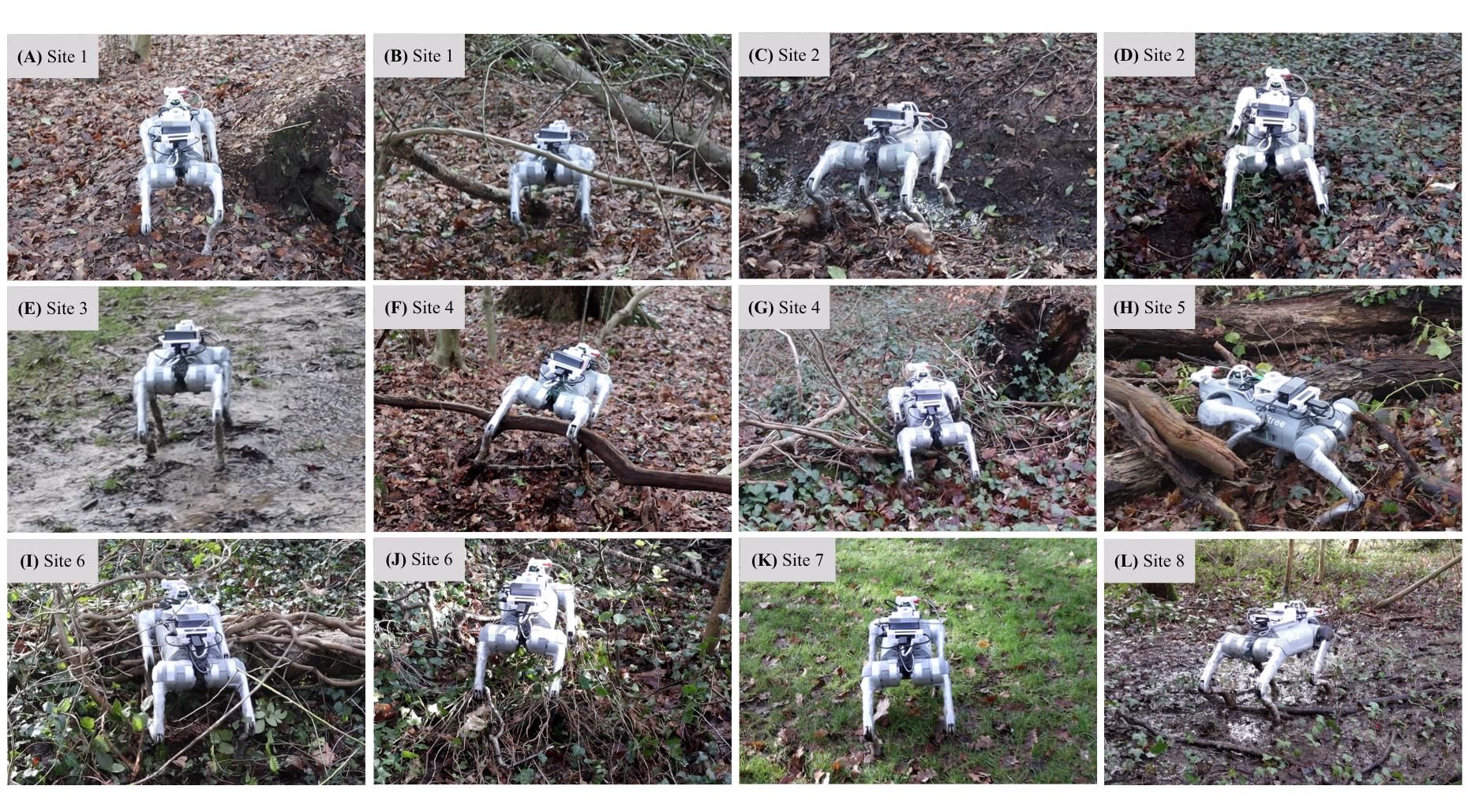}
    \caption{Representative third-person views of the Unitree Go2 Edu traversing diverse physical and visual terrain features across the eight test sites. Encountered environments include: (A) rocky incline, (B) overhanging vegetation, (C) ditch, (D) soft soil and vines, (E) mud and sand, (F) fallen branch, (G) piled branches and vines, (H) piled timber, (I) exposed roots and fallen tree, (J) dense ground vines, (K) short grass, and (L) waterlogged bog.}
    \label{fig:terrains}
\end{figure*}

\subsection{Dataset Sequences}
The ForEnt dataset comprises 11 continuous sequences recorded across 8 unstructured forest test sites, yielding a total traversed distance of 1.7~km (summarized in Table \ref{tab:dataset_sequences}). Traversed forest environments include isolated and piled fallen timber, ivy, vines, roots, mud, ditches, rocky inclines, ground depressions, thick leaf litter, waterlogged soil, and overhanging vegetation (see Fig.~\ref{fig:terrains}). In our field experiments, sequences that terminated prematurely due to environment-induced torque safety shutdowns or structural sensor-mount damage are annotated as such and intentionally retained to document severe mobility breakdowns. The retention of these terminal events accounts for the multiple sequences recorded at high-risk sites. 

\subsection{Entrapment Types} \label{subsec:Entrapment_Types}
During our field trials, four primary physical failure mechanisms were observed that resulted in entrapment:

\begin{compactitem}
  \item \textbf{Entanglement}: Vegetation entangled one or more joints during the swing phase, most often a KFE joint,  restricting the movement of the affected swing leg and consequently destabilizing the body. These events were most common in dense understorey, where vines were hidden beneath leaf litter.

  \item \textbf{Trip}: The collision between a swing leg and a rigid obstacle, typically the raised edge of a fallen log, causing the robot to become immobilized or lose balance.

  \item \textbf{Soft Soil}: The feet of the robot leg in the stance phase sank into deformable terrain, typically loose soil or waterlogged ground, increasing drag during the swing phase and causing a locomotion stall.

  \item \textbf{Suspension}: The robot's torso rested upon an elevated obstacle, typically a raised log or thick branch, partially lifting the legs off the ground and resulting in a loss of foot traction.
\end{compactitem}

To illustrate the multi-modal data captured during these events, Fig.~\ref{fig:teaser} presents a sequential window of third-person images alongside the corresponding exteroceptive and proprioceptive sensor streams for two failure cases: vegetation entanglement and a rigid trip on a fallen branch.

  
\begin{table*}[t]
\centering
\caption{Overview of the ForEnt dataset sequences. Traversed distances vary depending on the navigable area of each test site. Multiple sequences are recorded at a single site when a terminal failure forced a full restart of the site; these aborted sequences are retained to demonstrate extreme mobility breakdowns. Recorded data sizes represent the raw ros2bag files. Entrapment mechanisms: Entanglement (Ent.), Trips, Soft Soil (Soil), and Suspension (Susp.).}
\label{tab:dataset_sequences}
\begin{tabular*}{\textwidth}{@{\extracolsep{\fill}} c c c c c c c c c c c c @{}}
\toprule
\multirow{2}{*}{\textbf{Test Site}} & \multirow{2}{*}{\textbf{Seq.}} & \multirow{2}{*}{\textbf{Distance (m)}} & \multicolumn{4}{c}{\textbf{Recorded Data}} & \multicolumn{5}{c}{\textbf{Entrapment Events}} \\
\cmidrule(lr){4-7} \cmidrule(lr){8-12}
& & & \textbf{Size (GB)} & \textbf{RGB-D} & \textbf{LiDAR} & \textbf{Proprioception} & \textbf{Ent.} & \textbf{Trip} & \textbf{Soil} & \textbf{Susp.} & \textbf{Total} \\
\midrule
1 & 1 & 199.74 & 16.37 & 9241 & 5809 & 203632 & 1 & 9 & 0 & 2 & 12 \\
\midrule
\multirow{2}{*}{2} & 2 & 165.25 & 10.56 & 5982 & 3607 & 144248 & 3 & 2 & 1 & 0 & 6 \\
 & 3 & 248.41 & 18.44 & 10448 & 6333 & 243191 & 2 & 3 & 3 & 0 & 8 \\
\midrule
3 & 4 & 227.32 & 14.49 & 8242 & 4990 & 172046 & 0 & 0 & 0 & 0 & 0 \\
\midrule
\multirow{2}{*}{4} & 5 & 245.24 & 18.87 & 10957 & 6636 & 140610 & 4 & 4 & 0 & 3 & 11 \\
 & 6 & 291.43 & 19.68 & 11439 & 6983 & 131271 & 4 & 7 & 0 & 0 & 11 \\
\midrule
5 & 7 & 143.67 & 12.04 & 6980 & 4218 & 93963 & 2 & 4 & 0 & 3 & 9 \\
\midrule
\multirow{2}{*}{6} & 8 & 12.36 & 1.86 & 1064 & 604 & 23933 & 2 & 0 & 0 & 0 & 2 \\
 & 9 & 73.99 & 7.97 & 4616 & 2787 & 66071 & 3 & 3 & 0 & 1 & 7 \\
\midrule
7 & 10 & 23.46 & 1.64 & 924 & 558 & 24726 & 0 & 0 & 0 & 0 & 0 \\
\midrule
8 & 11 & 67.57 & 7.42 & 4305 & 2565 & 62298 & 0 & 1 & 1 & 1 & 3 \\
\midrule
\midrule
\multicolumn{2}{c}{\textbf{Total}} & \textbf{1698.44} & \textbf{129.34} & \textbf{74198} & \textbf{45090} & \textbf{1305989} & \textbf{21} & \textbf{33} & \textbf{5} & \textbf{10} & \textbf{69} \\
\bottomrule
\end{tabular*}
\end{table*}

\subsection{Annotation Protocol}
To ensure labeling consistency across the dataset, time-aligned labels were generated using a two-phase, semi-automated annotation pipeline. In the first phase, an automated pass identifies candidate entrapment windows by isolating continuous periods exceeding 3~s where the error between the commanded and estimated actual forward velocity ($\bar{v}_{x} - v_{x}$) exceeded 0.3~m/s. This threshold was selected via trial-and-error to robustly detect entrapment events, while excluding state estimation noise and temporary slowdowns over uneven terrain. In the second phase, a human annotator reviewed the synchronized third-person video and sensor logs for each candidate window. False positives (e.g., nominal stopping or turning) were discarded. For verified mobility failures, the annotator manually tagged the event as one of the four previously defined failure mechanisms (see Sec.~\ref{subsec:Entrapment_Types}) and categorized it using the following three-tier severity system based on the required recovery action:

\begin{compactitem}
    \item \textbf{Level 1 (Teleoperated Recovery)}: Forward motion stops for over 3~s despite continuous velocity commands, requiring the operator to recover the robot via manual teleoperation without physical contact.
    
    \item \textbf{Level 2 (Physical Intervention)}: The robot remains stuck after 10--15~s of Level 1 recovery attempts, or it falls over, requiring the operator to physically lift and reposition the robot on the path.
    
    \item \textbf{Level 3 (System Failure)}: The robot experiences actuator torque safety shutdowns or structural damage to sensor mounts. The platform becomes inoperable, requiring the sequence to be aborted for a hard reboot or physical repair. 
\end{compactitem}

\begin{figure}
    \centering
    \includegraphics[width=\linewidth]{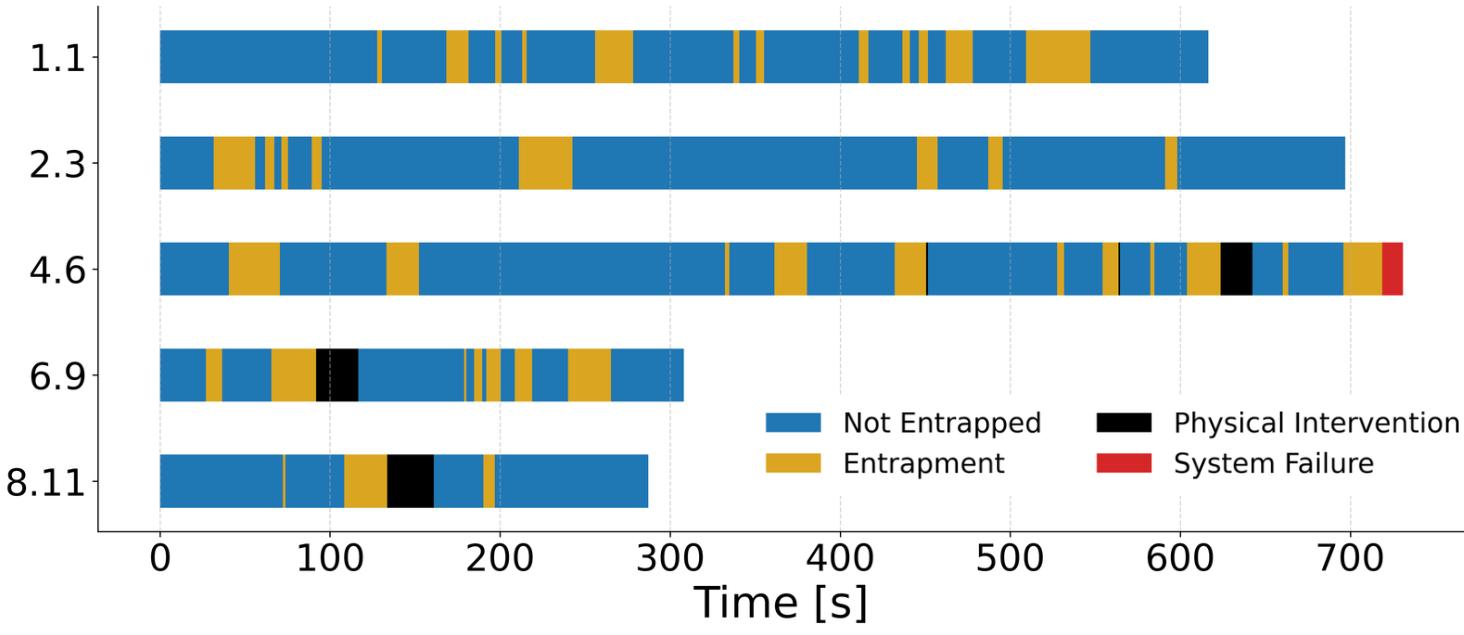}
    \caption{Temporal distribution of entrapment across representative dataset sequences (labeled as site index prefix and sequence index suffix, separated by a period). The timelines illustrate periods of nominal locomotion (blue) interrupted by mobility failures of varying severity: Level 1 entrapments requiring teleoperated recovery (yellow), Level 2 events necessitating physical intervention (black), and Level 3 terminal system failures (red).}
    \label{fig:label_seq}
    \vspace{-3mm}
\end{figure}

Every annotated event in the ForEnt dataset comprises the time interval of when the entrapment occurred, failure mechanism, and severity level. To illustrate the labeled sequence, Fig. \ref{fig:label_seq} visualizes the temporal distribution of entrapment and their severity across a representative sample of five sequences selected at random from the dataset. 

\subsection{Dataset Format}
\begin{figure}
    \centering
    \includegraphics[width=\linewidth]{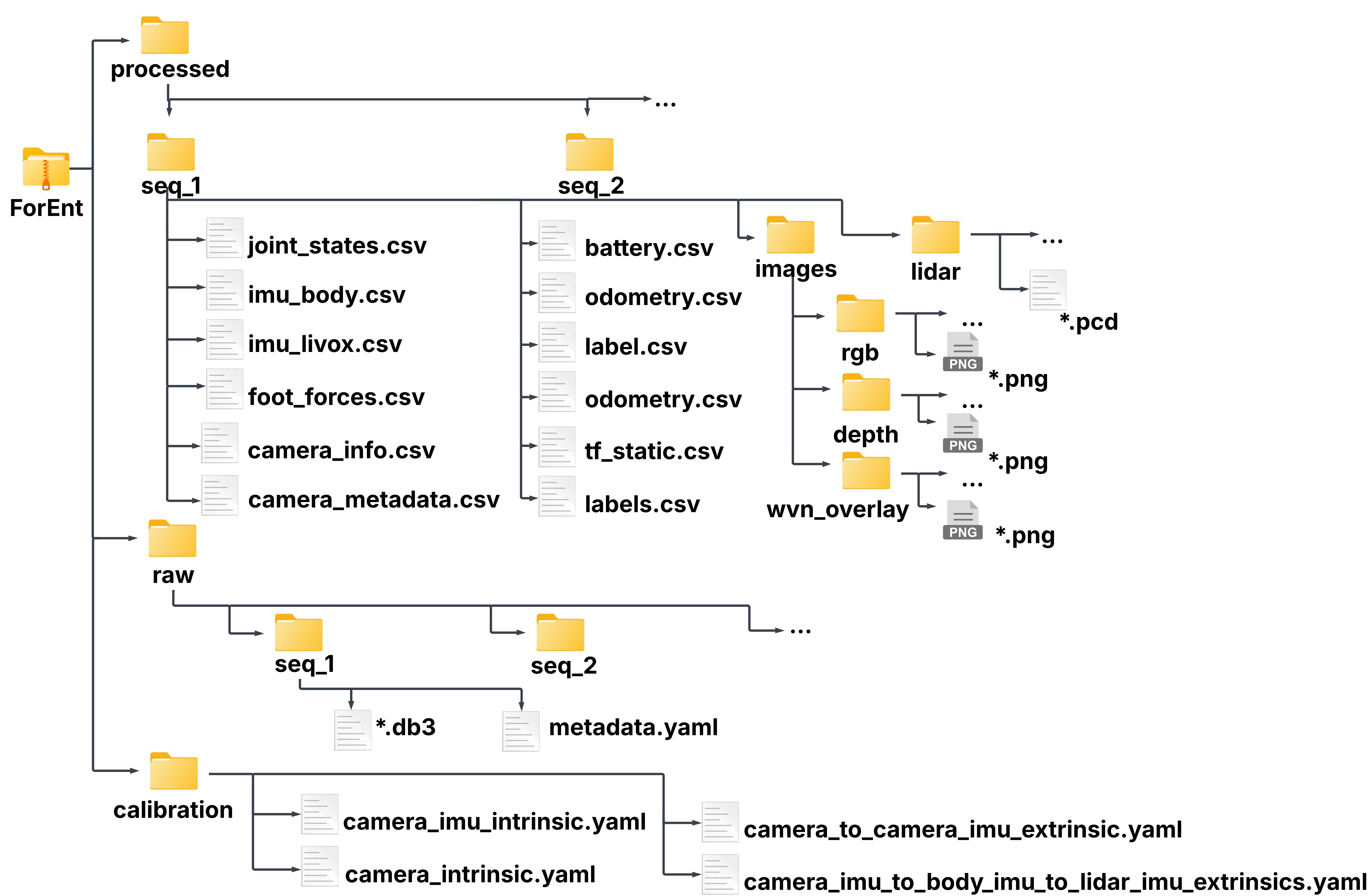}
    \caption{Directory structure of the ForEnt dataset. The repository is partitioned into three main directories: \texttt{raw} containing the original continuous ros2bag files, \texttt{processed} containing the extracted and time-synchronized multimodal sequence data (images, point clouds, and proprioceptive logs), and \texttt{calibration} containing the intrinsic and extrinsic sensor parameters.}
    \label{fig:dataset_structure}
    \vspace{-3mm}
\end{figure}

The ForEnt dataset is available on Zenodo \cite{Anonymous2025ForestEntrapmentDataset} and is organized into two primary subdirectories and includes relevant calibration parameters and processing scripts (see directory structure in Fig.~\ref{fig:dataset_structure}). Within both primary directories, data is further organized by sequence number (e.g., \texttt{raw/seq\_1/}). The \texttt{raw/} subdirectory contains the continuous sensor streams recorded as standard ros2bag files. Custom message definitions from the Unitree SDK and Livox drivers are provided to enable bag playback, and temporal alignment across modalities is maintained via the hardware-synchronized \texttt{header.stamp} fields. For use outside of the ROS~2 environment, the \texttt{processed/} subdirectory contains extracted and time-synchronized sequences. Within these site-specific folders, proprioceptive data, state estimates, and labels are stored as \texttt{.csv} files, RGB-D frames as \texttt{.png} images, and LiDAR scans as point cloud files (\texttt{.pcd}). Time-aligned third-person video recordings (\texttt{.mp4}) of the trials are also included. Finally, static intrinsic and extrinsic calibration parameters are provided alongside Bash (\texttt{.sh}) scripts to install required Ubuntu dependencies and execute data processing and extraction scripts.

%% file: Sections/Evaluation.tex
\section{Evaluation}
We demonstrate the utility of the ForEnt dataset by evaluating representative algorithms across two core autonomous navigation tasks: visual traversability estimation and proprioceptive entanglement detection.

\begin{figure*}
    \centering
    \includegraphics[width=\textwidth]{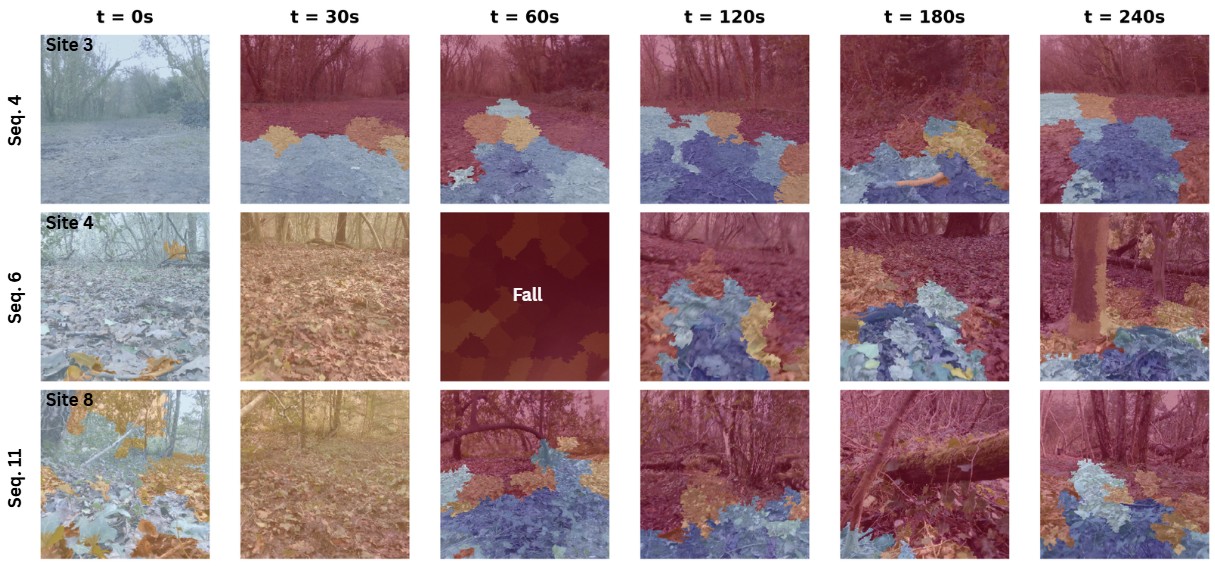}
    \caption{Examples of traversability estimation overlays generated by the WVN during training using DINO-ViT \cite{frey23fast}. The color map indicates predicted traversability, ranging from traversable (blue) to non-traversable (red), with columns denoting the elapsed time of the sequence. The model outputs non-traversable predictions for both rigid obstacles and visually novel areas that the robot has not yet seen. In Seq. 4 and Seq. 6, the model segments traversable paths from the surrounding environment. In Seq. 11, the model overestimates the traversability of waterlogged soil due to its visual similarity to solid dirt.}
    \label{fig:wvn_overlay}
\end{figure*}

\subsection{Visual Traversability Estimation}
ForEnt's suitability for learning-based perception is assessed by evaluating the Wild Visual Navigation (WVN) framework \cite{frey23fast}. We compare the performance of WVN using four visual feature extractors: DINO-ViT, ResNet-50 \cite{he2016resnet}, EfficientNet-B4 \cite{tan2019efficientnet}, and SIFT \cite{lowe2004sift}. The weights of the feature extractors are frozen; the downstream Multi-Layer Perceptron (MLP) is trained during the evaluation.

For each recorded sequence in ForEnt, the WVN MLP is trained on the first half of the sequence, and subsequently evaluated on the remaining half, for the binary classification of traversable versus non-traversable regions. To generate self-supervised ground-truth labels for training and evaluation, we define the mean squared velocity error as $v_{error} = 0.5((\bar{v}_{x} - v_{x})^2 + (\bar{v}_{y} - v_{y})^2)$, where $\bar{v}$ and $v$ represent the commanded and actual velocities in the $x$ and $y$ axes. The error is smoothed with a Kalman filter, sigmoid-activated, and binarized with a threshold of 0.1 to define the discrete traversability classes (for details see \cite{frey23fast}).

While the full WVN pipeline utilizes dynamic thresholding on the false positive rate for online deployment, our offline evaluation employs a fixed traversability decision threshold of 0.5, providing a stable, standardized baseline for sequence-level binary classification. Algorithm performance is quantified by comparing the predicted traversability against the binarized velocity error labels. To account for training variance, the classification accuracy was averaged across three random weight initializations. The final sequence-level results, excluding trajectories shorter than 30~m, are presented in Table \ref{tab:wvn_results}. A notable failure case emerges in Seq. 11, where all architectures exhibit a sharp decline in accuracy. As illustrated in Fig. \ref{fig:wvn_overlay}, this degradation stems from the visual ambiguity between waterlogged bog and firm soil, resulting in the models overestimating terrain traversability.

\begin{table}[htbp]
\centering
\caption{Sequence-Level Traversability Classification Accuracy ($\%$) Comparison Across Feature Extraction Architectures.}
\label{tab:wvn_results}
\begin{tabular*}{\columnwidth}{@{\extracolsep{\fill}} c cccc @{}}
\toprule
\textbf{Seq.} & \textbf{DINO-ViT} & \textbf{ResNet-50} & \textbf{EffNet-B4} & \textbf{SIFT} \\
\midrule
\textbf{1}  & 75.63 & 47.95 & 44.50 & 50.69 \\
\textbf{2}  & 79.07 & 67.12 & 75.42 & 40.70 \\
\textbf{3}  & 81.74 & 78.92 & 78.59 & 35.92 \\
\textbf{4}  & 63.09 & 85.55 & 92.60 & 86.58 \\
\textbf{5}  & 71.32 & 74.61 & 64.08 & 77.72 \\
\textbf{6}  & 84.10 & 77.13 & 78.88 & 88.86 \\
\textbf{7}  & 28.36 & 16.23 & 20.11 & 53.24 \\
\textbf{9}  & 43.86 & 47.97 & 49.17 & 51.76 \\
\textbf{11} & 14.07 & 19.89 & 11.16 & 10.51 \\
\bottomrule
\end{tabular*}
\vspace{-3mm}
\end{table}

\subsection{Proprioceptive Entanglement Detection}
To evaluate the utility of the ForEnt dataset for proprioceptive entanglement detection, we employ the Momentum-Based Observer (MBO) proposed in \cite{Yim2023EntanglementsIROS}. To the best of our knowledge, MBO is the only model specifically designed for detecting entanglement in legged robots through proprioceptive disturbance estimation. The observer estimates external joint torques by comparing measured and expected robot dynamics under the assumption that the robot’s torso behaves as an inertial reference frame during nominal locomotion. However, these assumptions are often violated in unstructured forest environments, where uneven terrain, vegetation interaction, and variable contact conditions induce base accelerations and non-stationary body motion.

\begin{figure}
    \centering
    \includegraphics[width=0.9\linewidth]{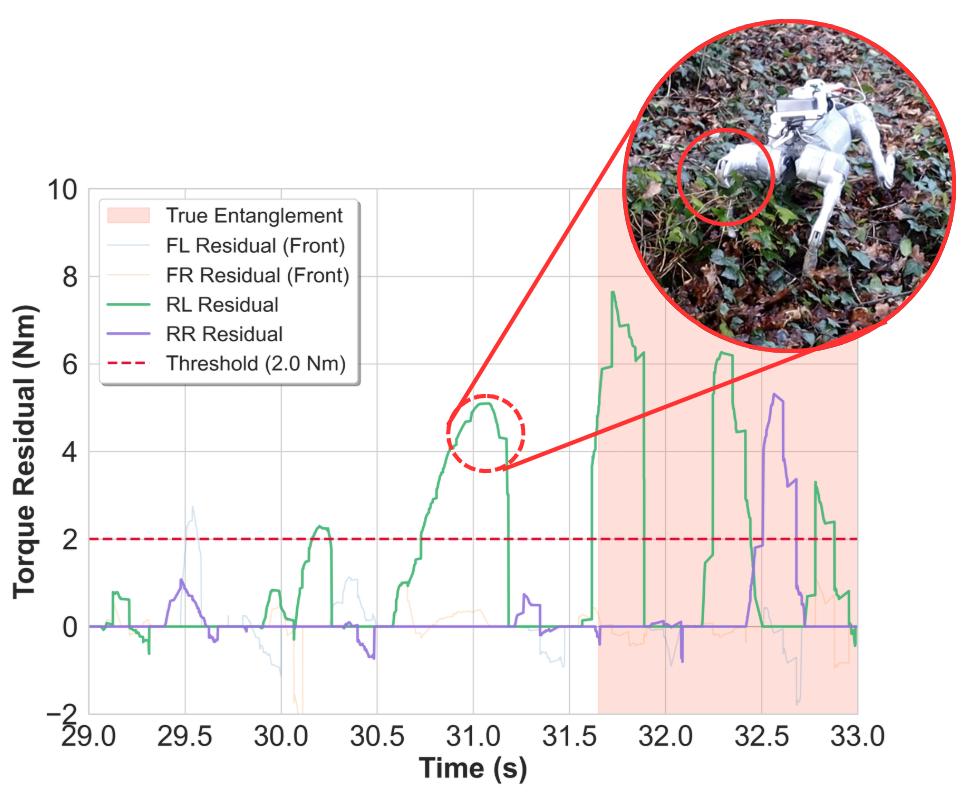}
    \caption{Residual momentum, defined as the difference between modeled and measured torque in the Momentum-Based Observer (MBO) \cite{Yim2023EntanglementsIROS}, for the rear-left (RL) Knee Flexion/Extension (KFE) joint during an entanglement in Site~2, Seq.~2. The dashed line indicates the 2~Nm decision threshold, and the red shaded region corresponds to the labelled entrapment interval.}
    \label{fig:mbo}
    \vspace{-3mm}
\end{figure}

To assess MBO under conditions consistent with its underlying assumptions, we first selected sequences from the ForEnt dataset in which the robot maintained approximately constant velocity and stable body posture. The observer was adapted for the Go2 by tuning the observer gain to $K_0=5$ to reduce sensitivity to stance-phase disturbances and setting the dry-friction parameter to $1.0$ (for details on parameters see \cite{Yim2023EntanglementsIROS}). As MBO assumes a fixed swing-stance duty cycle, which is not satisfied by reinforcement learning-based locomotion, stance phase was instead determined using onboard foot contact force measurements (threshold at $0.1$~N). Under these conditions, MBO successfully detected entanglement in sequences where its assumptions were approximately satisfied, as illustrated in Fig.~\ref{fig:mbo}. However, evaluating MBO across the complete ForEnt dataset yielded a poor true positive rate of 16.81\%. The reduced performance reflects limitations of the MBO model when applied to locomotion in forest terrain.

%% file: Sections/Conclusion.tex
\section{Conclusion}
We present ForEnt, a multi-modal dataset of quadruped robot entrapment events collected with the Go2 Edu platform in realistic forest environments. Across 11 continuous sequences, the dataset captures approximately 1.7~km of traversal and provides 69 annotated mobility failures. These entrapment events are documented with time-synchronized RGB-D imagery, 3D LiDAR scans, high-frequency proprioceptive measurements, and third-person video.

The recorded sequences highlight the physical limitations of blind reinforcement learning locomotion in unstructured forests, demonstrating that critical mobility breakdowns frequently occur from the interaction with terrain features such as large fallen logs and compliant vegetation. By providing a dedicated, failure-centric benchmark, the ForEnt dataset facilitates the reproducible offline training and validation of navigation algorithms without risking hardware damage.